\documentclass[letterpaper, 10 pt, conference]{ieeeconf}  

\IEEEoverridecommandlockouts                              

\usepackage{epsfig}
\usepackage{graphicx}
\usepackage{amsmath}
\usepackage{amssymb}

\usepackage[table]{xcolor}
\usepackage{subcaption}
\usepackage{bbm}
\usepackage{dsfont} 
\usepackage{soul}

\newcommand{\ie}[1][ ]{{\em i.\thinspace{}e\@.{}},#1}
\newcommand{\eg}[1][ ]{{\em e.\thinspace{}g\@.{}},#1}

\newcommand{\todo}[1]{{\textcolor{blue}{[TODO: #1]}}}
\newcommand{\wonttodo}[1]{}

\newcommand{\refsec}[1]{Sec.~\ref{sec:#1}}
\newcommand{\reffig}[1]{Fig.~\ref{fig:#1}}

\newcommand{\reftab}[1]{Table~\ref{tab:#1}}

\newcommand{\myparagraph}[1]{\vspace{3pt}\noindent{\bf #1}}

\definecolor{col1}{RGB}{213, 229, 255}
\definecolor{col2}{RGB}{128, 179, 255}
\definecolor{col3}{RGB}{181, 232, 151}
\definecolor{col4}{RGB}{235, 155, 221}

\title{\LARGE \bf
Occupancy Flow Fields for Motion Forecasting in Autonomous Driving
}

\author{Reza Mahjourian$^{*1}$, Jinkyu Kim$^{*2}$, Yuning Chai$^{1}$, Mingxing Tan$^{3}$, Ben Sapp$^{1}$, Dragomir Anguelov$^{1}$
\thanks{$^{1}$Waymo. {\tt\small rezama@waymo.com}}%
\thanks{$^{2}$Korea University.  Work done while at Waymo.}%
\thanks{$^{3}$Google Brain.}%
\thanks{$^{*}$Equal contribution.}%
}

\begin{document}

\maketitle
\thispagestyle{empty}
\pagestyle{empty}

\begin{abstract}
We propose Occupancy Flow Fields, a new representation for motion forecasting of multiple agents, an important task in autonomous driving.
Our representation is a spatio-temporal grid with each grid cell containing both the probability of the cell being occupied by any agent, and a two-dimensional flow vector representing the direction and magnitude of the motion in that cell. 
Our method successfully mitigates shortcomings of the two most commonly-used representations for motion forecasting: trajectory sets and occupancy grids.  Although occupancy grids efficiently represent the probabilistic location of many agents jointly, they do not capture agent motion and lose the agent identities.  To this end, we propose a deep learning architecture that generates Occupancy Flow Fields with the help of a new flow trace loss that establishes consistency between the occupancy and flow predictions.  We demonstrate the effectiveness of our approach using three metrics on occupancy prediction, motion estimation, and agent ID recovery. In addition, we introduce the problem of predicting speculative agents, which are currently-occluded agents that may appear in the future through dis-occlusion or by entering the field of view. We report experimental results on a large in-house autonomous driving dataset and the public INTERACTION dataset, and show that our model outperforms state-of-the-art models.
\end{abstract}

\section{Introduction}
\label{sec:intro}

\begin{figure}[htb!]
    \centering
        \includegraphics[width=\linewidth]{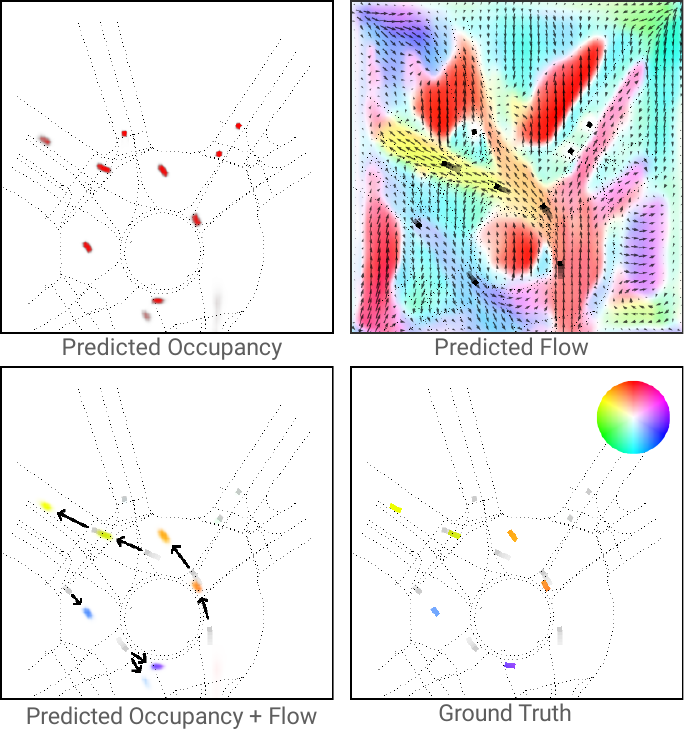}
    \caption{(Best viewed in color) Our model predicts occupancy (\textbf{top-left}) and flow (\textbf{top-right}) at discrete timesteps in the future.  Combining occupancy and flow produces a rich occupancy representation that captures the direction and magnitude of motion as well (\textbf{bottom-left}).  Notice the agent near the bottom and the two prominent predictions corresponding to staying in the roundabout and exiting.  The ground truth (\textbf{bottom-right}) contains the color wheel that maps motion direction and magnitude to different colors.  The gray traces in the bottom images show the past state of the agents.  All views are at $t + 1.5 s$.  The sample scene is from the Interaction dataset~\cite{interactiondataset}.}
    \label{fig:teaser}
    \vspace{-10pt}
\end{figure}

In this work, we tackle the problem of predicting the future location and motion of other vehicles and pedestrians from the view of an autonomous vehicle (AV).  To represent future locations, we adopt the notion of occupancy grids~\cite{thrun1996integrating} from the robotics literature.  An occupancy grid $O_t$, at a particular timestep $t$, can be represented by a single-channel gray-scale image with dimensions $h \times w$ where each pixel corresponds to a particular grid cell in the map, and the pixel value represents the probability that any part of any agent occupies that grid cell.  To represent future motion, we draw inspiration from optical flow.  A flow field $F_t$ at time $t$ can be represented by a two-channel image with dimensions $h \times w$ where each pixel holds a two-dimensional motion vector $(\Delta x, \Delta y)$ for the corresponding grid cell.  \reffig{teaser} illustrates sample occupancy and flow predictions vs. ground truth on a sample scene.

Motion forecasting is an essential component of planning in a multi-agent environment, and of particular interest for autonomous driving.  Modeling the uncertain future as a distribution over a compact set of trajectories per agent is a very popular choice~\cite{interactiondataset, buhet2021plop, chai2019multipath, chang2019argoverse, cui2019multimodal, phan2020covernet}. On the other hand, occupancy grid-based methods~\cite{bansal2018chauffeurnet, casas2021mp3, hong2019rules, jain2020discrete} provide some significant advantages: the non-parametric output captures a richer class of future distributions, incorporates shape and identity uncertainty, and models the joint probability of the existence of any agent in a spatio-temporal cell (rather than independent marginal probabilities per agent).  
While these observations make occupancy grids an attractive choice, occupancy grid methods have the disadvantage that agent identity is lost, and there is no obvious way to extract motion from the grids (which are a snapshot of a time interval), making it unclear how to interpolate at finer time granularity, and impossible to know the velocity of the agents.

This motivates \textit{Occupancy Flow Fields} which extend standard occupancy grids with flow fields.  By augmenting the output with flow estimates, we are able to trace occupancy from far-future grid locations back to current time locations by following the sequence of predicted flow vectors.  This gives us a way to recover the most-likely agent identity for any future grid cell.  

Another advantage for producing flow predictions is that it allows an occupancy model to capture future behavior with fewer ``key frames'', since flow predictions can be used to warp/morph occupancy at any continuous point in time.  Since our flow formulation captures multiple travel directions for each agent, the morphing process will lead to a \emph{conservative} expansion of occupancy from the last known occupancy of agents.  Therefore, the morphed occupancy can be safely used by a planning algorithm that minimizes co-location with predicted occupancy.

Given the current state-of-the-art in real-time perception, the track quality available to a motion forecasting system can be limited. We may lose sight of its tracked agents because of occlusions or increased distance.  More importantly, new agents may appear through dis-occlusion or otherwise entering the AV's field of view.  Reasoning about the location and velocity of these so-called {\em speculative agents} is critical for safe and effective autonomous driving. 
Trajectory prediction models in the literature are formulated only for agents that have already been detected and tracked, and cannot handle agents that may come out of occluded areas.

In summary, our contributions are as follows:

\noindent \textbf{Occupancy Flow Fields}: We introduce a model that predicts both occupancy and flow in a spatio-temporal grid.  This representation allows us to predict a non-parametric distribution of future occupancy as well as velocity of agents.

\noindent \textbf{Flow-Traced Occupancy}: We use the chain of flow predictions over multiple timesteps to trace predicted occupancies at any future timestep all the way back to the current observations.  Adding a loss term based on traced occupancy predictions forces the model to establish consistency between flow and occupancy predictions, leading to significant improvements in metrics.  At inference time, tracing flow predictions allows us to recover the identity of the agent predicted to occupy any grid cell.

\noindent \textbf{Speculative Agents}: We introduce the problem of predicting occupancy and flow for speculative agents---an important task that to our knowledge is unexplored in the behavior modeling literature.

\section{Related Work}

\myparagraph{Motion Forecasting via Trajectories.}
In this motion forecasting representation, each modeled agent's future distribution is described by a set of trajectories, which are each a time sequence of state estimates.  This set may be predicted in a discriminative, feed-forward manner, via imitation learning~\cite{bansal2018chauffeurnet, casas2018intentnet, chai2019multipath, helbing1995social, luo2018fast, pellegrini2009you, phan2020covernet, sadeghian2018car, phan2020covernet, chang2019argoverse, cui2019multimodal}.  These models commonly predict trajectory likelihoods and sometimes Gaussian uncertainty parameters as well, giving rise to a full parametric probability distribution as output~\cite{mercat2020multi, chai2019multipath, buhet2021plop, salzmann2020trajectron++}. 
The trajectory representation has several disadvantages mentioned in the introduction.

\myparagraph{Motion Forecasting via Occupancy Grids.}
There are relatively fewer motion forecasting models that employ an occupancy grid representation. ChauffeurNet~\cite{bansal2018chauffeurnet} trains occupancy in a multi-task network to improve trajectory planning performance. DRF~\cite{jain2020discrete} predicts a sequence of occupancy residuals inspired by auto-regressive sequential prediction. Rules of the Road~\cite{hong2019rules} compares trajectory methods to occupancy grids by proposing a dynamic program to decode likely trajectories from occupancy under a simple motion model. Finally, contemporaneous with this work, MP3~\cite{casas2021mp3} proposes a concept similar to Flow Fields, termed Motion Fields:  They predict a set of forward motion vectors and associated probabilities per grid cell.  MP3 employs occupancy flow in the context of a planning task and does not offer direct analysis on the quality and performance of their motion forecasting method.  In this work we explore occupancy flow in detail and develops a class of metrics for directly evaluating it.  

\section{Method}

In this section, we define the occupancy flow problem, describe our model architecture and losses, and elaborate on how we trace flow predictions over time to establish consistency between flow and occupancy predictions.

\subsection{Representation}

As introduced in \refsec{intro}, the problem is to predict an occupancy flow field at future time $t$, given observations from the recent state of the agents.  Following the common convention in AV motion datasets~\cite{ettinger2021large, interactiondataset, chang2019argoverse}, we abstract the agents as two-dimensional rectangles in bird's-eye view (BEV), characterized by position, orientation, width, height, velocity, etc.  A detection and tracking pipeline extracts the sparse agent states over a number of timesteps from raw sensor readings.  A sequence of observations is split into past and future segments.  Our model receives the past agent states as sparse inputs and predicts dense future occupancy.  Ground-truth occupancy is generated by rendering the bird's-eye view rectangles for the detected agents at each timestep.

\subsection{Inputs}
\label{sec:inputs}

We abstract the problem inputs to be sparse environment and agent states as estimated by any detection and tracking system as follows:

\textbf{1. Past agent states}: Each agent at time $t$ is represented as a tuple ($p_t$, $\theta_t$, $w_t$, $l_t$, $v_t$, $a_t$), where $p_t = (x_t, y_t)$ denotes the agent's center position, $\theta_t$ denotes the orientation, $(w_t, l_t$) denotes the box width and length, $v_t$ denotes a two-dimensional velocity vector, and $a_t$ denotes a two-dimensional acceleration vector.  
The model receives $A_t$, the state of all agents at time $t$, for $t \in \{T_\text{input}, \dots, 0\}$.

\textbf{2. Road structure}: To receive information about the structure of the road lanes and other traffic objects, the model is given a set of points sampled uniformly from the line segments and curves representing the road elements.  Each sampled point is represented by a tuple ($p$, $u$) where $p = (x, y)$ denotes position and $u$ denotes the type of the underlying road element, which can be one of the following: crosswalk, speed bump, stop/yield sign, road edge boundary, parking line, dotted line, solid single/double line,  and solid double yellow line.

\textbf{3. State of traffic lights}: The model is also given the state of traffic lights for each lane at each input timestep.  The traffic light state of each traffic-controlled lane at time $t$ is represented by a tuple $(p_t, s_t)$ where $p_t = (x_t, y_t)$ is the position of a point placed at the end of the traffic-controlled lane, and $s_t$ is the light state, which is one of $\{$red, yellow, green, unknown$\}$.

\subsection{Occupancy Flow Prediction}
\label{sec:problem}

Occupancy Flow Fields can be represented as two quantities: an occupancy grid $O_t$ and a flow field $F_t$, both with spatial dimensions $h \times w$. Each cell in the grid corresponds to a particular BEV grid cell in the map.  Each cell $(x, y)$ in the occupancy grid $O_t$ contains a value in the range $[0, 1]$ representing the probability that any part of any agent box overlaps with that grid cell at time $t$.  Each cell $(x, y)$ in the flow field $F_t$ contains a two-dimensional vector $(\Delta x, \Delta y)$ that specifies the motion of any agent whose box occupies that grid cell at time $t$.  
Vehicle and pedestrian behavior differ significantly, so we output separate occupancy and flow predictions for different agent classes $\mathcal{K}$.  More specifically, we predict occupancy grids $O_t = (O^{V}_{t}, O^{P}_{t})$ and flow fields $F_t = (F^{V}_{t}, F^{P}_{t})$ for vehicles and pedestrians, $\forall t \in \{1, \dots, T_\text{pred}\}$.

\begin{figure}[htb!]
    \vspace{5pt}
    \centering
        \includegraphics[width=\linewidth]{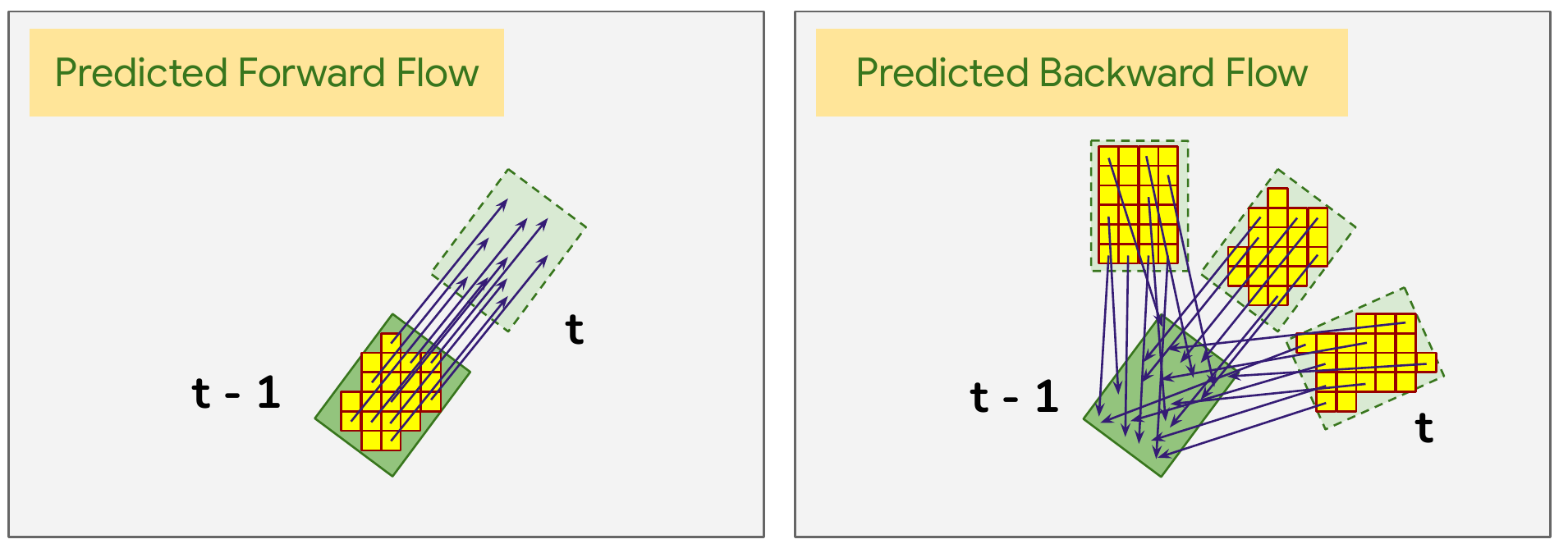}
    \caption{Hypothetical flow predictions for a single agent are illustrated in a single forward flow field (\textbf{left}) and a single backward flow field (\textbf{right}).  Each flow vector is stored in the grid cell (yellow square) at its base.  With forward flow, each grid cell predicts its future location, while with backward flow each grid cell predicts its past location.  Note that the agent boundaries drawn at time $t$ are just illustration aids and not predicted by the model.  Unlike forward flow, a single backward flow field can represent multiple next destinations for any current occupancy, making it more effective for motion forecasting.  Note that backward flow vectors are meaningful and predicted on every currently-unoccupied cell in the map (not shown), but forward flow is only meaningful on current occupancies.  Moreover, since each backward flow vector pulls occupancy from a single source, predicted backward flow for multiple agents is also inherently free of collisions--which indicate inconsistent predictions and are undesired in motion forecasting systems.}
    \label{fig:forward_backward_flow}
    \vspace{-5pt}
\end{figure}

\myparagraph{Flow Formulation:} We model the motion of agents with \emph{backward flow} (see \reffig{forward_backward_flow}).  Ground-truth flow vectors between times $t$ and $t - 1$ are placed in the grid at time $t$ and point to the original position of that grid cell at time $t - 1$.  More specifically, flow ground truth is constructed as $\tilde{F}_t(x, y) = (x, y)_{t-1} - (x, y)_{t}$, where $(x, y)_{t-1}$ denotes the coordinates at time $t - 1$ of the same agent part that occupies $(x, y)$ at $t$.  The magnitude of the flow vectors is in grid cell (think pixel) units.

Note that backward flow still models the forward motion of agents; it just represents where each grid cell comes from in the previous timestep rather than representing where each grid cell moves to in the next timestep.  Therefore backward flow can capture multiple futures for individual agents using a single flow field per timestep.  On the other hand, capturing multiple futures with forward flow requires predicting multiple flow vectors per cell and their associated probabilities, which would increase latency, memory requirement, and complexity of the model.

\subsection{Speculative Occupancy Flow Prediction}

The problem described in \refsec{problem} is to predict future occupancy and flow of the agents that have been observed at any of the past timesteps.  The speculative prediction model has the same inputs and the same output representation as the main model.  However, the task is to predict occupancy and motion of agents that have not been observed in the past, yet appear in the future, \eg a vehicle appearing on the edges of the model's field-of-view, or a pedestrian exiting a sensor-occluded region.  For this problem, we train the same model with alternative labels that reflect occupancy and flow of agents known to exist in the future but missing in past timesteps.  

\subsection{Model}

\begin{figure*}
    \centering
        \includegraphics[width=0.95\linewidth]{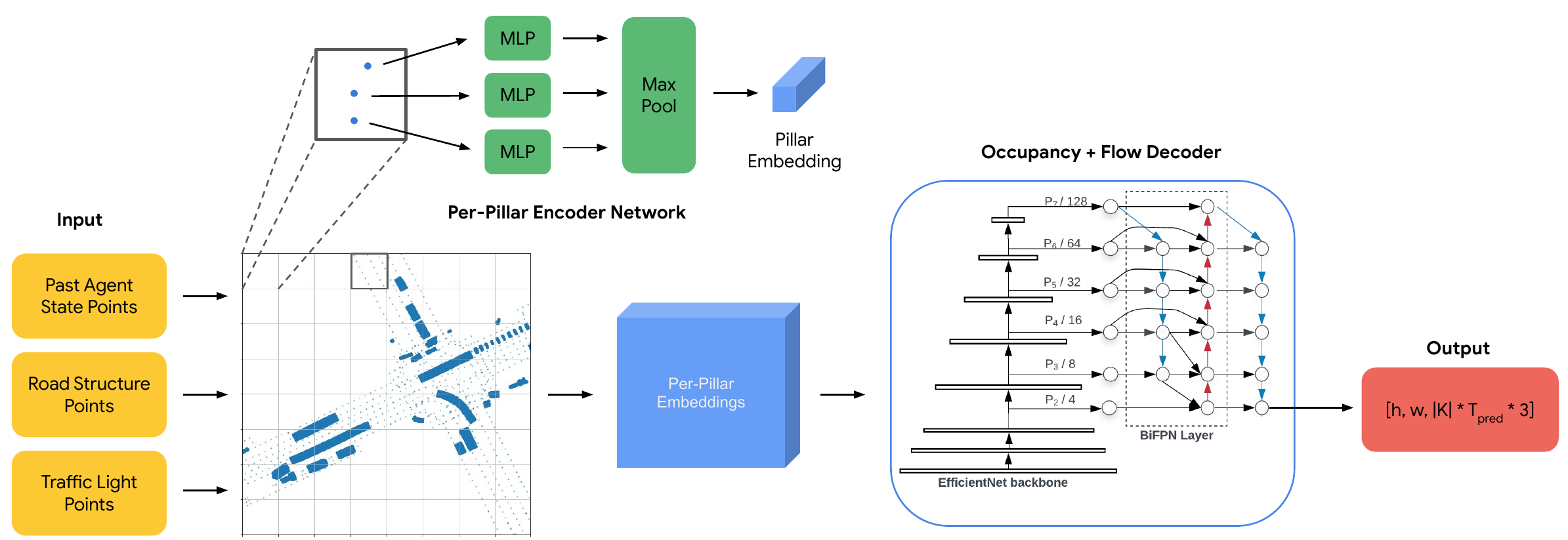}
    \caption{Our model architecture consists of PointPillars-inspired encoder~\cite{lang2019pointpillars, kim2021stopnet} and a decoder based on EfficientDet~\cite{tan2020efficientdet}.}
    \label{fig:arch}
    \vspace{-10pt}
\end{figure*}

\reffig{arch} shows the overall architecture of our model, which consists of an encoder and a decoder:

\myparagraph{Encoder:} the first stage receives all three types of input points and processes them with a PointPillars-inspired encoder ~\cite{lang2019pointpillars, kim2021stopnet}.  The traffic light and road points are placed directly into the grid.  The agent states $A_t$ at each input timestep $t \in \{T_\text{input}, \dots, 0\}$ are encoded by uniformly sampling a fixed-size grid of points from the interior of each agent's BEV box and placing those points with associated agent state attributes (\refsec{inputs}, including a one-hot encoding of time $t$) into the grid (visualized in \reffig{arch}).  Each pillar outputs an embedding for all the points contained in it.

\myparagraph{Decoder:} The second stage receives the per-pillar embeddings as input, and produces per-grid-cell occupancy and flow predictions.  The decoder network is based on EfficientDet~\cite{tan2020efficientdet}: it employs EfficientNet as the backbone to process the per-pillar embeddings into feature maps ($P_2$, ...$P_7$), where $P_i$ is downsampled by $2^i$ from inputs.  These multi-scale features are then fused in a bidirectional manner using a BiFPN network.  Then, the highest-resolution feature map $P_2$ is used to regress occupancy and flow predictions for all agent classes $\mathcal{K}$ over all $T_\text{pred}$ timesteps.  More specifically, the decoder outputs a vector of size $|\mathcal{K}| \times T_\text{pred} \times 3$ for each grid cell, in order to simultaneously predict occupancy ($|\mathcal{K}| \times T_\text{pred}$ channels) and flow ($|\mathcal{K}| \times T_\text{pred} \times 2$ channels).

\subsection{Losses}

The model is trained with supervised occupancy and flow field losses, and a novel self-supervised Flow Trace loss. The occupancy loss is a binary logistic cross-entropy per grid cell between the predicted $O$ and groundtruth occupancy $\tilde{O}$ aggregated over all spatio-temporal cells as
\\
\begin{equation}
    \mathcal{L}_O = \sum_{t=1}^{T_\text{pred}} \sum_{x=0}^{w-1} \sum_{y=0}^{h-1} \mathcal{H}(O_t(x, y), \tilde{O}_t(x, y))
\end{equation}
\\
where $\mathcal{H}$ denotes the cross-entropy function.

The flow loss is an L1-norm regression loss with respect to the ground-truth flow $\tilde{F}$, weighted by $\tilde{O}$ as
\\
\begin{equation}
    \mathcal{L}_F = \sum_{t=1}^{T_\text{pred}} \sum_{x=0}^{w-1} \sum_{y=0}^{h-1} \Big\lVert F_t(x, y) - \tilde{F}_t(x, y) \Big\rVert_1   \tilde{O}_t(x, y).
\end{equation}

\subsubsection{Flow Trace Loss}
\label{sec:warp}

\begin{figure*}
\vspace{0.5em}
\centering
    \includegraphics[width=0.95\linewidth]{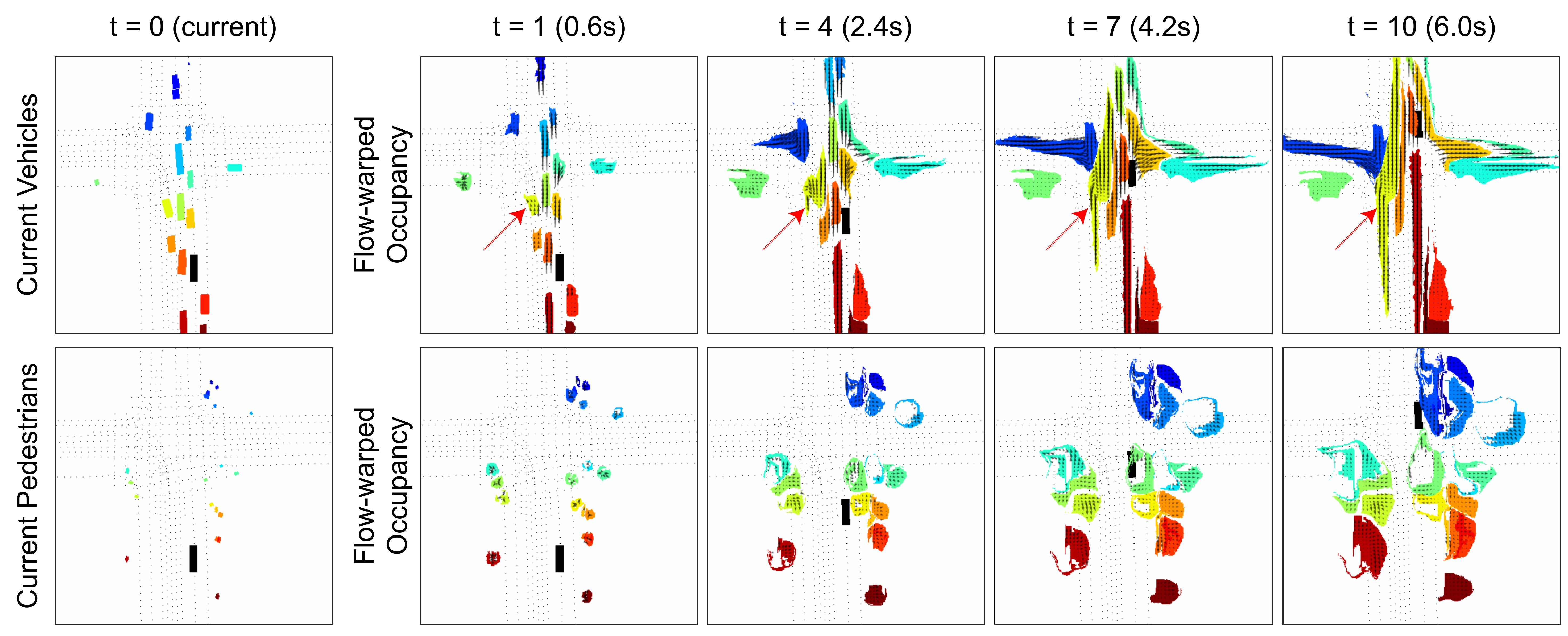}
    \vspace{-5pt}
    \caption{Flow Traces. \textbf{Top}: Current vehicles occupancy $\tilde{O}_{v_0}$ followed by recursive construction of flow-warped occupancies $\mathcal{W}_{v_t}$.  Each agent ID has been mapped to a different color.  The black box shows the ground-truth location of the AV.  At each timestep, flow predictions are used to expand the potential reachable region for each agent, independently of the occupancy likelihoods.  Since the backward flow field can pull any occupied grid cell to multiple future locations, each successive application of the warping process likely increases the reachable region.  For example, notice how the vehicle highlighted by the red arrow is predicted to either go straight, or perform a U-turn.  The flow-warped occupancies are used in loss functions at training time.  At inference time, they can be used to recover agent identities for predicted occupancies.  \textbf{Bottom}: The same process applied to the pedestrians in the same scene.  Note that the predicted flow vectors have been inverted to make it easier to study the predicted motions.}
\label{fig:trace}
\vspace{-15pt}
\end{figure*}

As discussed in \refsec{problem}, the backward flow vectors at time $t$ point to the previous location of corresponding grid cells at time $t - 1$.  Consider the current occupancy grid for vehicles $O_0$.  Note that $O_0$ is not a prediction, but can be constructed directly from inputs.  We can warp the current occupancy $O_0$ according to the first flow prediction $F_1$ for $t = 1$ to obtain a grid of all possible future occupancies of the current agents at $t = 1$.  
We recursively apply this warping process as
\\
\begin{equation}
    \mathcal{W}_t = F_t \circ \mathcal{W}_{t - 1}
\end{equation}
\\
where $\mathcal{W}_t$ denotes the flow-warped occupancy at $t$ and $\mathcal{W}_0 = O_0$.  Therefore, computing $\mathcal{W}_T$ for the final timestep applies a chain of all flow predictions $F_t, \forall t \in 1 \ldots T$ 
to compute all potential future locations for the current occupancies at $O_0$.  \reffig{trace} visualizes this process in a sample scene.  Note that we are using backward flow fields to roll out the current occupancies forward in time.  But, thanks to backward flow vectors, there are never any overlaps/conflicts on the origin of cells in $\mathcal{W}_t$.  Moreover, backward flow vectors never move current occupancies out of the grid.

The flow warping process does not use any occupancy predictions.  However, we multiply each $\mathcal{W}_t$ with its corresponding occupancy prediction $O_t$, and require that the result matches the ground truth occupancy $\tilde{O}_t$ using the loss term
\begin{equation}
    \mathcal{L}_{\mathcal{W}} = \sum_{t=1}^{T_\text{pred}} \sum_{x=0}^{w-1} \sum_{y=0}^{h-1} \mathcal{H}(\mathcal{W}_t(x,y) O_t(x,y), \tilde{O}_t(x,y)).
\end{equation}

The final loss is defined as
\\
\begin{equation}
    \mathcal{L} = \sum_{\ensuremath{\mathcal{K}}} 
    \frac{1}{hwT_\text{pred}} (\lambda_{O} \mathcal{L}_{O} + \lambda_{F} \mathcal{L}_{F}  + \lambda_\mathcal{W} \mathcal{L}_{\mathcal{W}})
\end{equation}
\\
where $\lambda_{O}, \lambda_{F}, \lambda_\mathcal{W}$ are coefficients, and $\ensuremath{\mathcal{K}}$ contains all agent classes, \ie vehicles and pedestrians in our case.

\subsection{Recovering Agent IDs Using Flow Traces}

The flow traces discussed as a loss in \refsec{warp} can also be used at inference time to assign agent IDs to predicted occupancies.
If we augment the current occupancy grid $O_0$ with the ID of the origin agent for every grid cell, the warping process spreads the agent ID across the flow vectors as well.  In this setup, $\mathcal{W}_t$ contains flow-warped occupancies with per-cell ID attributes, from which we can directly read the agent ID.  This ID points to the agent that \emph{could} occupy this grid cell at time $t$ according to the flow predictions $F_1, \dots, F_t$.  
Construction of flow traces and thereby the ID recovery process is very fast with time complexity $O(h \cdot w \cdot T_\text{pred})$.

\section{Experiments}
\label{sec:exp}


\subsection{Datasets}

\myparagraph{Crowds Dataset.}
This dataset is a revision of the Waymo Open Motion Dataset~\cite{ettinger2021large} focused on crowded scenes.  It contains 10.5 million training and 2.8 million test examples spanning over 500 hours of real-world driving in several urban areas across the US.  Dynamic scene entities are computed from LiDAR and camera data similar to existing works in the literature~\cite{fairfield2011traffic, yang2018hdnet}.  All scenarios contain at least 20 dynamic agents.  This dataset contains sensor readings at 5Hz.  We produce predictions for six seconds into the future, given one second of observations.

\myparagraph{Interaction~\cite{interactiondataset}.}
Interaction is a publicly-available dataset with sensor readings at 10Hz.  We use 432k examples for training and 108k examples for test.  We produce predictions for three seconds into the future, given half a second of observations.


\subsection{Training Setup}

We produce predictions for 30 future timesteps using observations from the past 5 timesteps $(T_\text{input} = -4$).  For both datasets, each prediction timestep aggregates occupancy and flow from 3 timesteps in the dataset.  Therefore $T_\text{pred} = 10$ captures future occupancy over all 30 timesteps.  The encoder uses $80 \times 80$ pillars, each mapping to a $1m \times 1m$ area of the world. The occupancy and flow outputs have a resolution of $h \times w = 400 \times 400$ cells, covering the same $80m \times 80m$ area.  Loss coefficients are set to $\lambda_O = \lambda_\mathcal{W} = 1000, \lambda_F = 1$ to roughly balance the magnitude of different losses.  The model is trained from scratch using the Adam optimizer with a learning rate of $0.02$ and batch size of $4$.


\subsection{Metrics}

\myparagraph{Occupancy Metrics}: We employ evaluation metrics used for binary segmentation~\cite{chen2017deeplab}: Area under the Curve (AUC) and Soft Intersection over Union (Soft-IoU)~\cite{mattyus2017deeproadmapper}.
\textbf{AUC} computes $\text{AUC}(O^\mathcal{K}_t, \tilde{O}^\mathcal{K}_t)$ for agent class $\mathcal{K}$ (vehicle/pedestrian) using a linearly-spaced set of thresholds in $[0, 1]$ to compute pairs of precision and recall values and estimate the area under the PR-curve.
\textbf{Soft-IoU} measures the area of overlap for each agent class $\mathcal{K}$ using the Soft IoU metric as 
\begin{equation}
\text{Soft-IoU}(O^\mathcal{K}_t, \tilde{O}^\mathcal{K}_t) = \frac{\sum_{x,y} O^\mathcal{K}_t  \cdot \tilde{O}^\mathcal{K}_t}{\sum_{x,y} O^\mathcal{K}_t + \tilde{O}^\mathcal{K}_t - O^\mathcal{K}_t  \cdot \tilde{O}^\mathcal{K}_t}
\end{equation}
\\
where arguments $(x, y)$ and have been omitted for brevity.

\myparagraph{Flow Metrics}:  The following metrics measure the accuracy of flow predictions:
\textbf{EPE} computes the mean End-Point Error L2 distance $\Big\lVert F^\mathcal{K}_t(x, y) - \tilde{F}^\mathcal{K}_t(x, y)\Big\rVert_2$ where $\tilde{O}^\mathcal{K}_{t}(x, y) \neq 0$.
\textbf{ID Recall} measures the percentage of correctly-recalled IDs for each ground-truth occupancy grid $\tilde{O}^\mathcal{K}_t$ as
\begin{equation}
\frac{
\sum_{x,y} \mathds{1} [\text{ID}(\mathcal{W}^\mathcal{K}_t) = \text{ID}(\tilde{O}^\mathcal{K}_t)] . \mathds{1}[\tilde{O}^\mathcal{K}_t \ne 0]
}{
\sum_{x,y} \mathds{1}\big[ \tilde{O}^\mathcal{K}_t \ne 0 \big]
}
\end{equation}
\\
where $\mathds{1}[]$ denotes the indicator function.

\myparagraph{Combined Metrics}: These metrics require both flow and occupancy predictions to be accurate:
\textbf{Flow-Traced (FT) AUC} measures $\text{AUC}(\mathcal{W}^\mathcal{K}_t  O^\mathcal{K}_t, \tilde{O}^\mathcal{K}_t)$.
\textbf{Flow-Traced (FT) IoU} measures $\text{Soft-IoU}(\mathcal{W}^\mathcal{K}_t  O^\mathcal{K}_t, \tilde{O}^\mathcal{K}_t)$.


\subsection{Results}

We report results for applying our method to three separate occupancy flow predictions tasks: 1) on the Crowds dataset, 2) for speculative objects in the Crowds dataset, and 3) on the Interaction dataset.

\begin{figure*}[htb!]
  \centering
   \includegraphics[width=.9\linewidth]{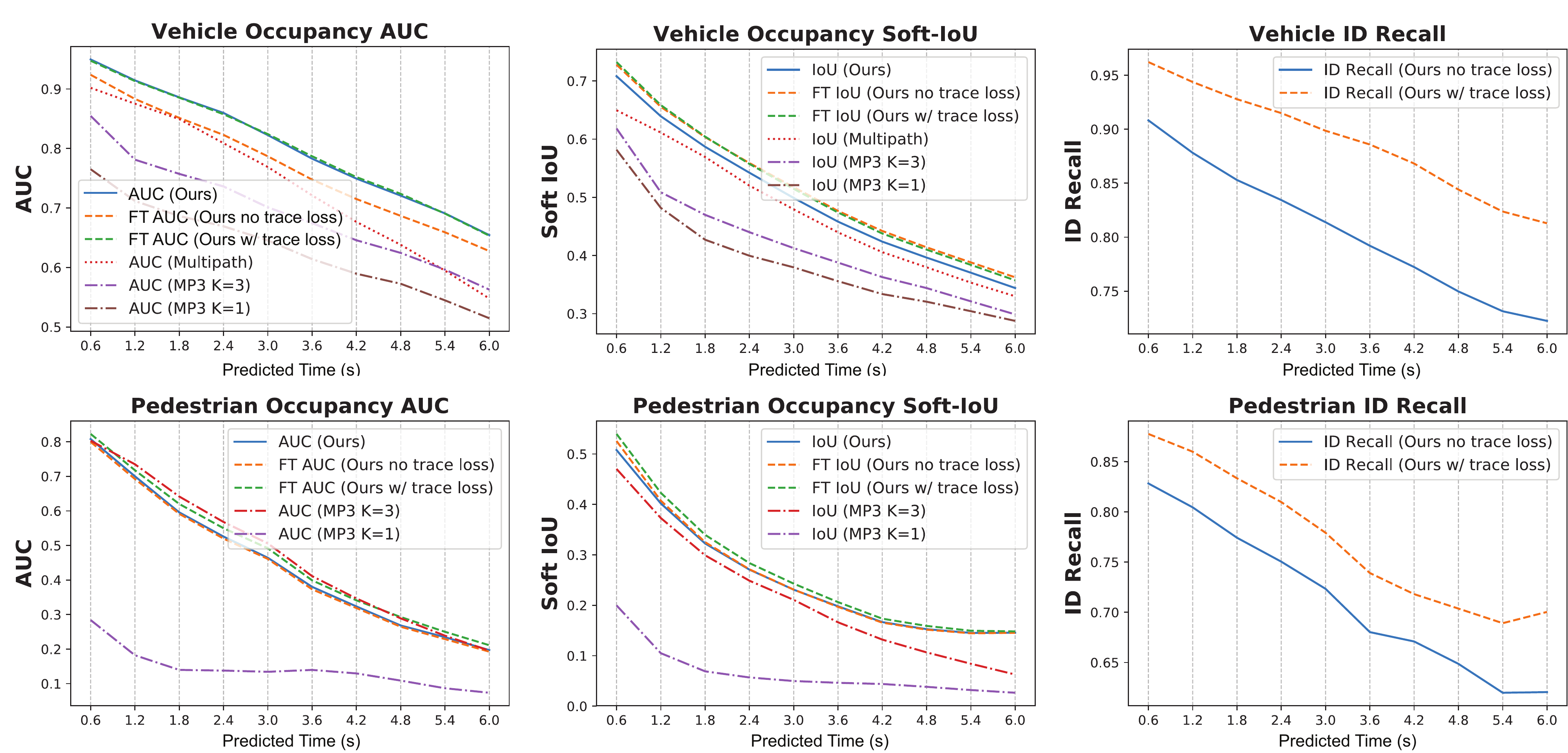}
\vspace{-5pt}
\caption{Occupancy and flow metrics separately for vehicles and pedestrians on the Crowds dataset.  The plots compare occupancy and ID recall metrics from two models trained with and without the flow-trace loss with MP3~\cite{casas2021mp3}.  
For vehicles, we also compare the models to occupancy grids generated from MultiPath~\cite{chai2019multipath}, a trajectory prediction model.}
\label{fig:plot_waymo}
\end{figure*}

\begin{figure}[htb!]
    \centering
        \includegraphics[width=0.9\linewidth]{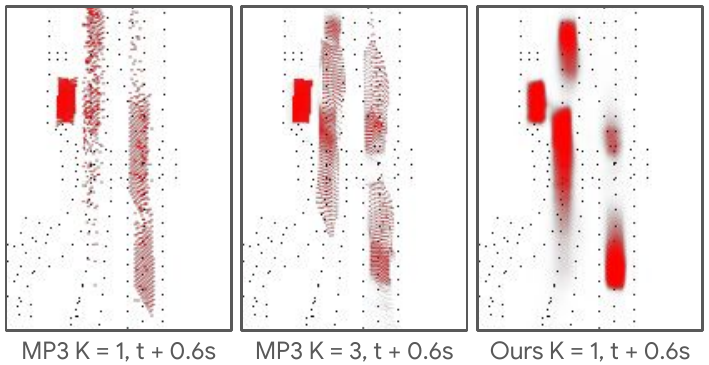}
    \caption{Predictions by MP3 (with $K=1$, $K=3$ flow fields) and our model ($K=1$) on a sample scene.  Unlike MP3, our backward forward flow fields have the capacity to disperse predicted occupancy in noisy datasets.}
    \label{fig:compare-mp3}
    \vspace{-15pt}
\end{figure}

\begin{figure*}[htb!]
    \vspace{0.5em}
    \centering
        \includegraphics[width=\linewidth]{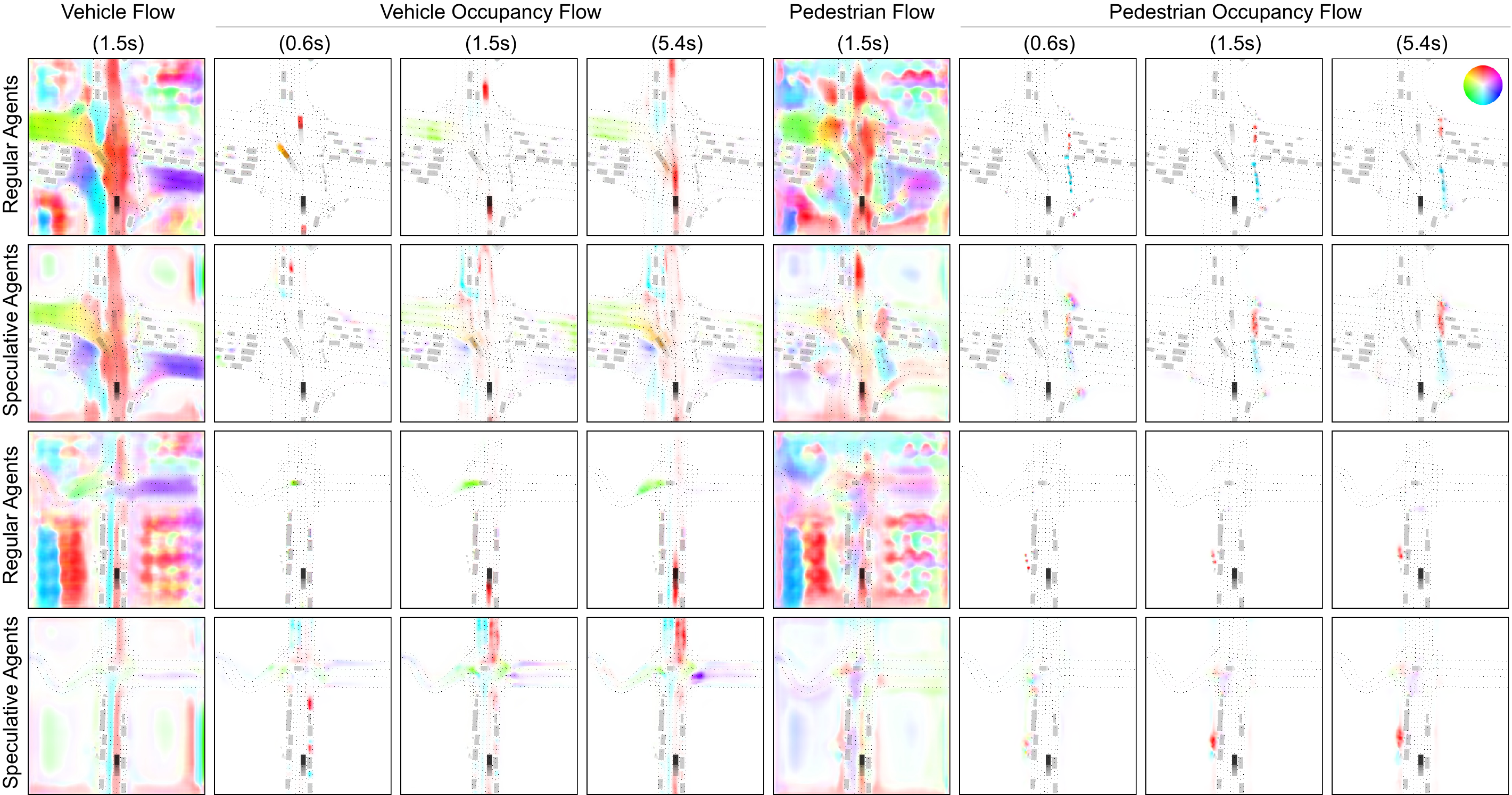}
    \caption{Regular and speculative predictions on two sample scenes (\textbf{top two} and \textbf{bottom two} rows) from the Crowds dataset. The \textbf{left four columns} display predictions for vehicles, and the \textbf{right four columns} show pedestrians.
    For each scene, a single flow prediction ($F_t$) and three combined flow and occupancy predictions ($F_t . O_t$) are shown.  Gray boxes show the recent state of input agents and the clouds visualize predicted occupancy and flow.  Regular occupancy is predicted on the path of moving agents.  Speculative occupancy is predicted in regions which might contain agents currently hidden from the AV (black box near bottom).}
    \label{fig:vis_waymo_and_spec}
    \vspace{-0.5em}
\end{figure*}

\begin{figure*}[htb!]
  \centering
  \includegraphics[width=.8\linewidth]{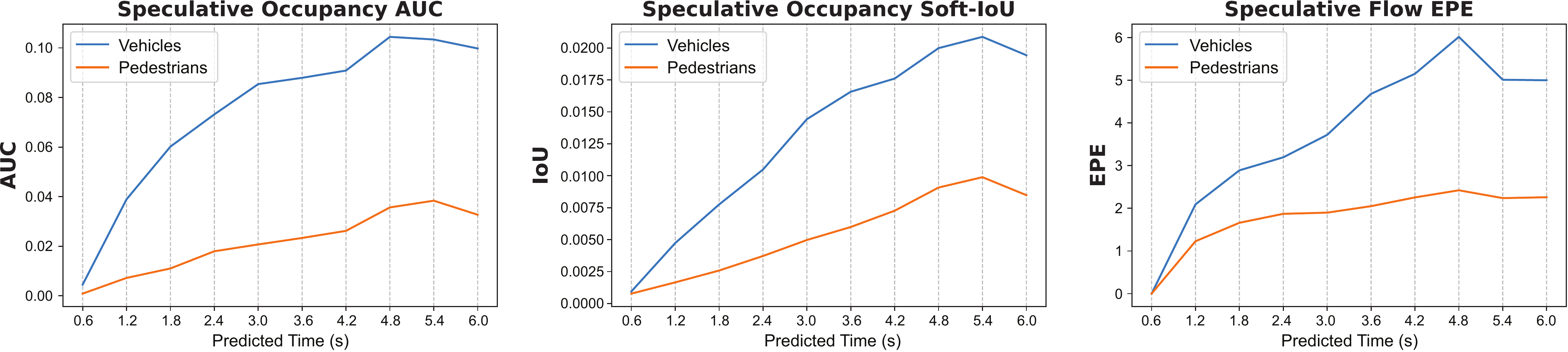}
\caption{Occupancy (AUC, Soft-IOU) and flow (EPE) metrics for the speculative model on the Crowds dataset.}
\label{fig:plot_waymo_spec}
\vspace{-10pt}
\end{figure*}

\begin{table*}[!ht]
  \centering
  \resizebox{0.88\textwidth}{!}{
  \begin{tabular}{|c||c|c|c|c|c|c|c|c|c|c|c|c|}
  \multicolumn{1}{c}{} & \multicolumn{6}{|c|}{\cellcolor{col1}Ours without trace loss} & \multicolumn{6}{c|}{\cellcolor{col3}Ours with trace loss} \\
  \hline
  Time &
  \multicolumn{2}{c|}{\cellcolor{col1}Occupancy} &  \cellcolor{col1}Flow & \cellcolor{col1}ID & \multicolumn{2}{c|}{\cellcolor{col1}Flow-Traced Occupancy} & \multicolumn{2}{c|}{\cellcolor{col3}Occupancy} &  \cellcolor{col3}Flow & \cellcolor{col3}ID & \multicolumn{2}{c|}{\cellcolor{col3}Flow-Traced Occupancy} \\
  (sec) &
  \cellcolor{col1}AUC & \cellcolor{col1}IoU &  \cellcolor{col1}EPE & \cellcolor{col1}Recall & \cellcolor{col1}AUC & \cellcolor{col1}IoU & 
  \cellcolor{col3}AUC & \cellcolor{col3}IoU &  \cellcolor{col3}EPE & \cellcolor{col3}Recall & \cellcolor{col3}AUC & \cellcolor{col3}IoU \\
  \hline
  0.3  & 0.938 & 0.802 & 0.439 & 0.899 & 0.920 & 0.796 & 0.939 & 0.802 & 0.458 & 0.950 & 0.941 & 0.817 \\
  0.6  & 0.898 & 0.720 & 0.456 & 0.841 & 0.876 & 0.708 & 0.899 & 0.723 & 0.536 & 0.915 & 0.896 & 0.734 \\
  0.9  & 0.853 & 0.636 & 0.528 & 0.800 & 0.825 & 0.626 & 0.860 & 0.641 & 0.632 & 0.886 & 0.851 & 0.652 \\
  1.2  & 0.796 & 0.551 & 0.591 & 0.762 & 0.764 & 0.544 & 0.801 & 0.557 & 0.736 & 0.859 & 0.796 & 0.563 \\
  1.5  & 0.717 & 0.461 & 0.667 & 0.724 & 0.687 & 0.457 & 0.730 & 0.472 & 0.797 & 0.825 & 0.723 & 0.476 \\
  1.8  & 0.651 & 0.389 & 0.722 & 0.698 & 0.620 & 0.388 & 0.659 & 0.396 & 0.820 & 0.807 & 0.652 & 0.401 \\
  2.1  & 0.576 & 0.324 & 0.811 & 0.683 & 0.544 & 0.323 & 0.578 & 0.324 & 0.870 & 0.791 & 0.572 & 0.328 \\
  2.4  & 0.509 & 0.271 & 0.887 & 0.663 & 0.475 & 0.270 & 0.509 & 0.268 & 0.893 & 0.775 & 0.503 & 0.272 \\
  2.8  & 0.432 & 0.229 & 0.935 & 0.649 & 0.399 & 0.228 & 0.429 & 0.229 & 0.935 & 0.766 & 0.425 & 0.234 \\
  3.0  & 0.366 & 0.187 & 0.947 & 0.635 & 0.338 & 0.188 & 0.369 & 0.190 & 0.995 & 0.751 & 0.365 & 0.196 \\
  \hline
  \end{tabular}	  
  }
    \caption{Vehicle occupancy and flow metrics over time from our two models on the Interaction dataset.}
    \label{tab:results_inter}
    \vspace{-10pt}
\end{table*}


\begin{figure*}[htb!]
  \centering
  \includegraphics[width=\linewidth]{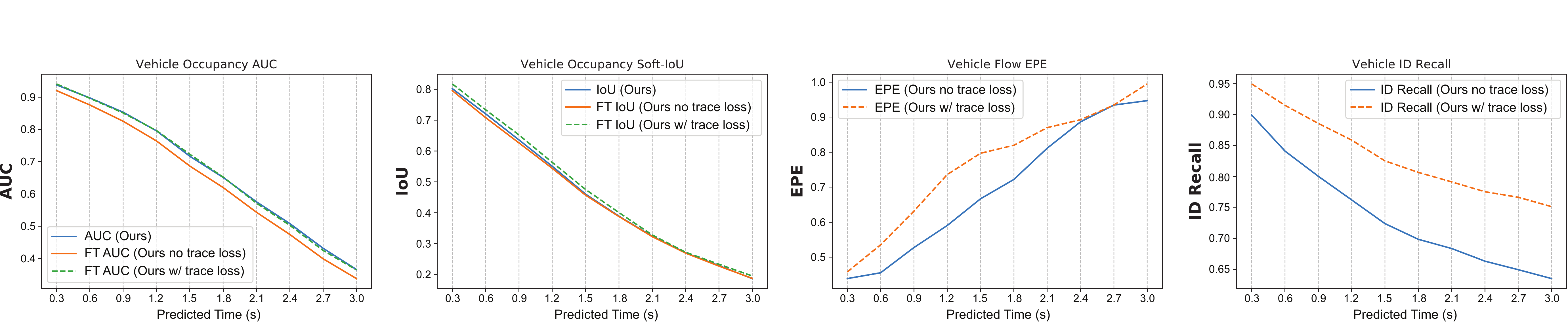}
\vspace{-10pt}
\caption{Occupancy and flow metrics on the Interaction dataset, which contains only vehicles.  The plots compare occupancy, flow, and ID recall metrics from two models trained with and without the flow-trace loss.  Training the model with the flow-trace loss leads to significant improvements.}
\label{fig:plot_inter}
\vspace{-10pt}
\end{figure*}

\reffig{plot_waymo} compares occupancy and flow metrics from two models trained with and without the flow-trace loss on the Crowds dataset.  For metrics that require the chain of flow predictions to be correct, \ie FT AUC, FT IoU, and ID recall, using the flow-trace loss leads to significant improvements.  \wonttodo{For vehicles at 6s, we observe an improvement of \todo{X\%} in flow-traced AUC, \todo{X\%} in flow-traced IoU, and \todo{X\%} in ID recall.}  Note that the flow-traced occupancy metrics tend to be lower than regular occupancy metrics, as they compare $\mathcal{W}_t O_t$ against ground truth and we have $\mathcal{W}_t O_t \leq O_t$.  In other words, the best the trace process in $\mathcal{W}_t$ can do is to reach all predicted occupancies and retain their intensities.  We observe that when training with the trace loss, the flow-traced occupancy metrics can reach the same level as the simpler occupancy metrics.  We notice that for pedestrians the flow-traced metrics can even surpass the simple metrics.  Since multiplying $\mathcal{W}_t$ can only lower the predicted occupancy values $O_t$, we conjecture that the improvement must be happening through $\mathcal{W}_t$ wiping out some predicted occupancies that are improbable according to flow predictions, and thereby improving the match with the ground truth.

We compare our occupancy prediction method with a state-of-the-art trajectory prediction model, MultiPath~\cite{chai2019multipath}, by converting its predictions to occupancy grids.  For a consistent comparison, we train the MultiPath backbone and decoder on the same feature maps obtained from our sparse encoder.  We convert the top 6 most-likely trajectories with likelihoods and Gaussian uncertainties to occupancy grids by rasterizing the predicted oriented agent boxes and convolving them with the two-dimensional Gaussian predicted for that timestep, weighted by the associated trajectory likelihood.  As \reffig{plot_waymo} shows, our rich non-parametric occupancy grid representation outperforms the trajectory model.

\reffig{plot_waymo} also compares our models with MP3~\cite{casas2021mp3}, which predicts a set of $K$ forward flow fields and associated probabilities.  Our single backward flow representation performs favorably against MP3 despite having a more compact representation.   We match MP3's performance with $K=3$ for pedestrians with just one flow field.  For vehicles, we outperform it even with $K = 3$.  MP3's flow representation is less compute- and memory-efficient.  For $K = 3$, each timestep needs 9 output channels compared to just 2 in our method.  We had to train separate MP3 models for each agent class to fit into the memory---giving it an advantages since our models predict all agent classes.


MP3 does not directly supervised occupancy, and its forward flow fields have a limited capacity in spreading occupancy in uncertain situations.  With $K=3$ fields, each currently-occupied pixel can move to at most three locations in the next timestep and touch at most 12 pixels with bilinear smoothing.  In \textbf{noisy datasets} where the motion of objects can be uncertain, MP3 struggles to cover all possible future locations of agents, and can produce predictions with agents disintegrating into disjoint pixels.  \reffig{compare-mp3} demonstrates this behavior.  
Moreover, our occupancy representation allows for modeling occupancy of \textbf{speculative agents}, a very important class of agents for AVs, which is not possible with MP3.

\reffig{plot_waymo_spec} shows occupancy and flow metrics for the speculative occupancy prediction task.  The metrics are worse in absolute values, since speculative prediction is a harder problem.  Note that unlike the main problem, speculative occupancy metrics improve over time, since speculative objects are harder to predict in near future than far future.  Most scenes have no agents disoccluding in the near future.  However, it is possible to anticipate potential disocclusions, \eg around the corners, with the forward motion of the AV or other agents. 
\reffig{vis_waymo_and_spec} visualizes regular and speculative occupancy flow predictions on two sample scenes.

\reffig{plot_inter} compares occupancy and flow metrics on the Interaction dataset from two models trained with and without the flow-trace loss.  We have included the metric values on this dataset in \reftab{results_inter} as well.  Again, we see significant improvements in metrics that depend on the chain of flow predictions.  Regular occupancy metrics slightly improve under the flow trace loss as well.  However the one-step flow accuracy metric regresses for the model trained with the trace loss, especially for early timesteps.  Since object detection and pose estimation models are often run independently from frame to frame, the AV datasets typically include very noisy readings.  Often even bounding box extents fluctuate between subsequent frames, which leads to noisy occupancy and flow labels in ground-truth.  The trace loss incentivizes the model to smooth out these fluctuations and learn the true behavior of the agents to be able to capture their long-term motion patterns, rather than trying to model the noise and temporal inaccuracies in the detection pipeline.

\section{Conclusions}

In this paper we proposed a motion forecasting model that predicts both occupancy and flow on a spatio-temporal grid, allowing us to predict not only the probabilistic location, but also the extents, motion, velocity and identity of agents (whether observed or speculative) in the future.  We also showed that our method for warping current occupancies based on flow predictions can improve our motion forecasting metrics.  Future work can explore gains from predicting speculative occupancies in an AV planning application.

\bibliographystyle{IEEEtran}
\bibliography{bibliography}

\begin{thebibliography}{10}
\providecommand{\url}[1]{#1}
\csname url@rmstyle\endcsname
\providecommand{\newblock}{\relax}
\providecommand{\bibinfo}[2]{#2}
\providecommand\BIBentrySTDinterwordspacing{\spaceskip=0pt\relax}
\providecommand\BIBentryALTinterwordstretchfactor{4}
\providecommand\BIBentryALTinterwordspacing{\spaceskip=\fontdimen2\font plus
\BIBentryALTinterwordstretchfactor\fontdimen3\font minus
  \fontdimen4\font\relax}
\providecommand\BIBforeignlanguage[2]{{%
\expandafter\ifx\csname l@#1\endcsname\relax
\typeout{** WARNING: IEEEtran.bst: No hyphenation pattern has been}%
\typeout{** loaded for the language `#1'. Using the pattern for}%
\typeout{** the default language instead.}%
\else
\language=\csname l@#1\endcsname
\fi
#2}}

\bibitem{interactiondataset}
W.~Zhan, L.~Sun, D.~Wang, H.~Shi, A.~Clausse, M.~Naumann, J.~K\"ummerle,
  H.~K\"onigshof, C.~Stiller, A.~de~La~Fortelle, and M.~Tomizuka, ``Interaction
  dataset: An international, adversarial and cooperative motion dataset in
  interactive driving scenarios with semantic maps,'' \emph{arXiv:1910.03088},
  2019.

\bibitem{thrun1996integrating}
S.~Thrun and A.~B{\"u}cken, ``Integrating grid-based and topological maps for
  mobile robot navigation,'' in \emph{Proceedings of the National Conference on
  Artificial Intelligence}, 1996, pp. 944--951.

\bibitem{buhet2021plop}
T.~Buhet, E.~Wirbel, A.~Bursuc, and X.~Perrotton, ``Plop: Probabilistic
  polynomial objects trajectory prediction for autonomous driving,'' in
  \emph{CoRL}, 2021, pp. 329--338.

\bibitem{chai2019multipath}
Y.~Chai, B.~Sapp, M.~Bansal, and D.~Anguelov, ``Multipath: Multiple
  probabilistic anchor trajectory hypotheses for behavior prediction,''
  \emph{CoRL}, 2019.

\bibitem{chang2019argoverse}
M.-F. Chang, J.~Lambert, P.~Sangkloy, J.~Singh, S.~Bak, A.~Hartnett, D.~Wang,
  P.~Carr, S.~Lucey, D.~Ramanan, \emph{et~al.}, ``Argoverse: 3d tracking and
  forecasting with rich maps,'' in \emph{CVPR}, 2019, pp. 8748--8757.

\bibitem{cui2019multimodal}
H.~Cui, V.~Radosavljevic, F.-C. Chou, T.-H. Lin, T.~Nguyen, T.-K. Huang,
  J.~Schneider, and N.~Djuric, ``Multimodal trajectory predictions for
  autonomous driving using deep convolutional networks,'' in \emph{ICRA}, 2019,
  pp. 2090--2096.

\bibitem{phan2020covernet}
T.~Phan-Minh, E.~C. Grigore, F.~A. Boulton, O.~Beijbom, and E.~M. Wolff,
  ``Covernet: Multimodal behavior prediction using trajectory sets,'' in
  \emph{CVPR}, 2020, pp. 14\,074--14\,083.

\bibitem{bansal2018chauffeurnet}
M.~Bansal, A.~Krizhevsky, and A.~Ogale, ``Chauffeurnet: Learning to drive by
  imitating the best and synthesizing the worst,'' \emph{RSS}, 2019.

\bibitem{casas2021mp3}
S.~Casas, A.~Sadat, and R.~Urtasun, ``Mp3: A unified model to map, perceive,
  predict and plan,'' in \emph{CVPR}, 2021, pp. 14\,403--14\,412.

\bibitem{hong2019rules}
J.~Hong, B.~Sapp, and J.~Philbin, ``Rules of the road: Predicting driving
  behavior with a convolutional model of semantic interactions,'' in
  \emph{CVPR}, 2019, pp. 8454--8462.

\bibitem{jain2020discrete}
A.~Jain, S.~Casas, R.~Liao, Y.~Xiong, S.~Feng, S.~Segal, and R.~Urtasun,
  ``Discrete residual flow for probabilistic pedestrian behavior prediction,''
  in \emph{CoRL}.\hskip 1em plus 0.5em minus 0.4em\relax PMLR, 2020, pp.
  407--419.

\bibitem{casas2018intentnet}
S.~Casas, W.~Luo, and R.~Urtasun, ``Intentnet: Learning to predict intention
  from raw sensor data,'' in \emph{CoRL}, 2018, pp. 947--956.

\bibitem{helbing1995social}
D.~Helbing and P.~Molnar, ``Social force model for pedestrian dynamics,''
  \emph{Physical review E}, vol.~51, no.~5, p. 4282, 1995.

\bibitem{luo2018fast}
W.~Luo, B.~Yang, and R.~Urtasun, ``Fast and furious: Real time end-to-end 3d
  detection, tracking and motion forecasting with a single convolutional net,''
  in \emph{CVPR}, 2018, pp. 3569--3577.

\bibitem{pellegrini2009you}
S.~Pellegrini, A.~Ess, K.~Schindler, and L.~Van~Gool, ``You'll never walk
  alone: Modeling social behavior for multi-target tracking,'' in \emph{ICCV},
  2009, pp. 261--268.

\bibitem{sadeghian2018car}
A.~Sadeghian, F.~Legros, M.~Voisin, R.~Vesel, A.~Alahi, and S.~Savarese,
  ``Car-net: Clairvoyant attentive recurrent network,'' in \emph{ECCV}, 2018,
  pp. 151--167.

\bibitem{mercat2020multi}
J.~Mercat, T.~Gilles, N.~El~Zoghby, G.~Sandou, D.~Beauvois, and G.~P. Gil,
  ``Multi-head attention for multi-modal joint vehicle motion forecasting,'' in
  \emph{ICRA}, 2020, pp. 9638--9644.

\bibitem{salzmann2020trajectron++}
T.~Salzmann, B.~Ivanovic, P.~Chakravarty, and M.~Pavone, ``Trajectron++:
  Multi-agent generative trajectory forecasting with heterogeneous data for
  control,'' \emph{arXiv preprint arXiv:2001.03093}, 2020.

\bibitem{ettinger2021large}
S.~Ettinger, S.~Cheng, B.~Caine, C.~Liu, H.~Zhao, S.~Pradhan, Y.~Chai, B.~Sapp,
  C.~R. Qi, Y.~Zhou, \emph{et~al.}, ``Large scale interactive motion
  forecasting for autonomous driving: The waymo open motion dataset,'' in
  \emph{ICCV}, 2021, pp. 9710--9719.

\bibitem{lang2019pointpillars}
A.~H. Lang, S.~Vora, H.~Caesar, L.~Zhou, J.~Yang, and O.~Beijbom,
  ``Pointpillars: Fast encoders for object detection from point clouds,'' in
  \emph{CVPR}, 2019, pp. 12\,697--12\,705.

\bibitem{kim2021stopnet}
J.~Kim, R.~Mahjourian, S.~Ettinger, M.~Bansal, B.~White, B.~Sapp, and
  D.~Anguelov, ``Stopnet: Scalable trajectory and occupancy predictionfor urban
  autonomous driving,'' in \emph{ICRA}, 2022.

\bibitem{tan2020efficientdet}
M.~Tan, R.~Pang, and Q.~Le, ``Efficientdet: Scalable and efficient object
  detection,'' in \emph{IEEE Conf. Comput. Vis. Pattern Recog.}, 2020.

\bibitem{fairfield2011traffic}
N.~Fairfield and C.~Urmson, ``Traffic light mapping and detection,'' in
  \emph{ICRA}, 2011, pp. 5421--5426.

\bibitem{yang2018hdnet}
B.~Yang, M.~Liang, and R.~Urtasun, ``Hdnet: Exploiting hd maps for 3d object
  detection,'' in \emph{CoRL}, 2018, pp. 146--155.

\bibitem{chen2017deeplab}
L.-C. Chen, G.~Papandreou, I.~Kokkinos, K.~Murphy, and A.~L. Yuille, ``Deeplab:
  Semantic image segmentation with deep convolutional nets, atrous convolution,
  and fully connected crfs,'' \emph{TPAMI}, vol.~40, no.~4, pp. 834--848, 2017.

\bibitem{mattyus2017deeproadmapper}
G.~M{\'a}ttyus, W.~Luo, and R.~Urtasun, ``Deeproadmapper: Extracting road
  topology from aerial images,'' in \emph{ICCV}, 2017, pp. 3438--3446.

\end{thebibliography}

\end{document}